\documentclass[journal,twoside,web]{ieeecolor}
\usepackage{generic}

\usepackage[utf8]{inputenc}
\usepackage[T1]{fontenc}
\usepackage{todonotes}
\usepackage{subcaption}
\usepackage{siunitx}
\usepackage[colorlinks=true, allcolors=blue]{hyperref}
\usepackage{float}
\restylefloat{table}
\usepackage{amsmath}
\usepackage{amsfonts}
\usepackage{pifont}
\usepackage{algorithmicx,algorithm}
\usepackage{booktabs}
\usepackage{algpseudocode}
\usepackage{multirow}
\usepackage{booktabs} 
\usepackage{makecell}
\usepackage{xcolor}
\usepackage{tikz}
\usepackage{stfloats} 
\usepackage{anyfontsize}
\usepackage{bbding}

\definecolor{lime}{HTML}{A6CE39}
\DeclareRobustCommand{\orcidicon}
{
    \begin{tikzpicture}
    \draw[lime, fill=lime] (0,0) circle [radius=0.16] 
    node[white] {{\fontfamily{qag}\selectfont \tiny ID}};    \draw[white, fill=white] (-0.0625,0.095) circle [radius=0.007];    
    \end{tikzpicture}
    \hspace{0mm}}
\foreach \x in {A, ..., Z}{%
    \expandafter\xdef\csname orcid\x\endcsname{\noexpand\href{https://orcid.org/\csname orcidauthor\x\endcsname}{\noexpand\orcidicon}}
}

\usepackage{color, colortbl}
\usepackage{diagbox}
\usepackage{graphicx}
\usepackage{caption}
\def\BibTeX{{\rm B\kern-.05em{\sc i\kern-.025em b}\kern-.08em
    T\kern-.1667em\lower.7ex\hbox{E}\kern-.125emX}}
\begin{document}
\title{\vspace*{-7pt}\fontsize{18pt}{18pt}\selectfont DIR-BHRNet: A Lightweight Network for Real-time Vision-based Multi-person Pose Estimation on Smartphones}

\author{\vspace*{-9pt}
Gongjin Lan\hspace{-1mm}\orcidA{}\hspace{-1mm}, Yu Wu, Qi Hao \Envelope\orcidB{}\hspace{-2mm}, \IEEEmembership{Member, IEEE}
\thanks{This work was supported in part by the National Natural Science Foundation of China (62261160654), the Shenzhen Fundamental Research Program (JCYJ20220818103006012), the Shenzhen Key Laboratory of Robotics and Computer Vision (ZDSYS20220330160557001), the GuangDong Basic and Applied Basic Research Foundation (2021A1515110641). (Corresponding author: Qi Hao.)}
\thanks{Gongjin Lan and Yu Wu are with the Department of Computer Science and Engineering, Southern University of Science and Technology, China}
\thanks{Qi Hao is with the Research Institute of Trustworthy Autonomous Systems, and the Department of Computer Science and Engineering, Southern University of Science and Technology, China. (e-mail: hao.q@sustech.edu.cn)}
\vspace*{-43pt}
}

\maketitle

\begin{abstract}
\renewcommand{\footnotesize}{\normalsize}
Human pose estimation (HPE), particularly multi-person pose estimation (MPPE), has been applied in many domains such as human-machine systems.
However, the current MPPE methods generally run on powerful GPU systems and take a lot of computational costs.
Real-time MPPE on mobile devices with low-performance computing is a challenging task.
In this paper, we propose a lightweight neural network, DIR-BHRNet, for real-time MPPE on smartphones.
In DIR-BHRNet, we design a novel lightweight convolutional module, Dense Inverted Residual (DIR), to improve accuracy by adding a depthwise convolution and a shortcut connection into the well-known Inverted Residual, and a novel efficient neural network structure, Balanced HRNet (BHRNet), to reduce computational costs by reconfiguring the proper number of convolutional blocks on each branch.
We evaluate DIR-BHRNet on the well-known COCO and CrowdPose datasets. 
The results show that DIR-BHRNet outperforms the state-of-the-art methods in terms of accuracy with a real-time computational cost.
Finally, we implement the DIR-BHRNet on the current mainstream Android smartphones, which perform more than 10 FPS.
The free-used executable file (Android 10), source code, and a video description of this work are publicly available on the page \footnote{\label{note1} \scriptsize \url{ https://github.com/wuyuuu/BalancedHighResolutionNetwork}}
 to facilitate the development of real-time MPPE on smartphones.
\end{abstract}

\vspace{-3pt}

\vspace{-11pt}
\section{Introduction}
\label{sec:introduction}
\vspace{-3pt}

Multi-person Pose Estimation (MPPE) is a crucial task in computer vision, which detects the position and orientation of the person by predicting the location of keypoints like hands, head, shoulders, elbows, wrists, hips, knees, and ankles. 
MPPE can improve human-computer interaction by enabling computers to recognize and understand human poses and movements.
It has been widely applied in various applications such as human-computer interaction, sports analysis, healthcare, robotics, and entertainment \cite{Lan2023vision}.
The current MPPE methods can be classified into two categories top-down and bottom-up methods.
Top-down methods generally perform real-time MPPE with a less computational cost for various numbers of persons (particularly crowds) since their computational costs significantly increase over the increasing number of detected persons.
Bottom-up methods \cite{newell2017associative,cao2019openpose} identify all joint keypoints and group them into personal instances, which perform a relatively stable computational cost MPPE for various numbers of persons.
In real-world applications, the stability of computational costs is crucial for real-time MPPE on smartphones.
Bottom-up methods generally outperform top-down methods for real-time MPPE of various numbers of persons with a stable computational cost on low-performance computing hardware such as smartphones.
However, most current bottom-up methods are computationally intensive and are typically executed on powerful GPU platforms.
Although current lightweight neural networks such as MobileNets \cite{sandler2018mobilenetv2} and ShuffleNets \cite{zhang2018shufflenet} have demonstrated advanced accuracy in MPPE, these take a lot of computational costs and generally need powerful GPU systems. 

\begin{figure*}[!ht] \centering
    \includegraphics[width=.98\textwidth,trim={5 150 760 63},clip]{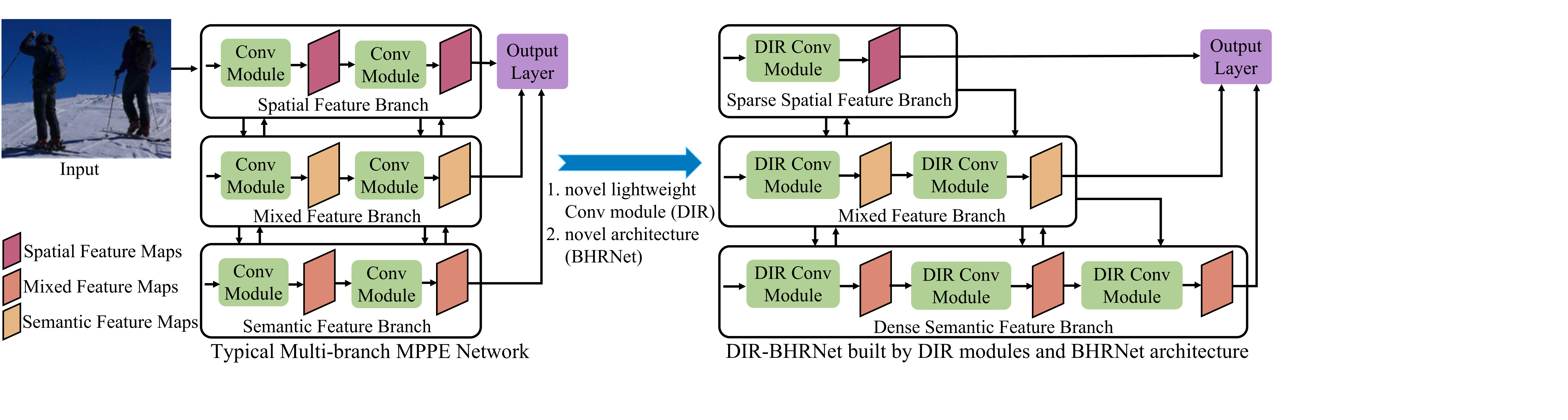} \vspace{-5pt}
    \caption{The comparison of the typical MPPE network and our DIR-BHRNet built by DIR modules and BHRNet architecture.}
    \label{fig:system-setup}
\end{figure*}

In this work, we propose a novel bottom-up method lightweight network, DIR-BHRNet, for real-time MPPE on smartphones.
Specifically, a novel lightweight convolutional block, Dense Inverted Residual (DIR), is proposed to improve accuracy.
In DIR, we add an extra depthwise convolution and a shortcut connection to extend the well-known Inverted Residual \cite{sandler2018mobilenetv2}, which extracts richer spatial features while maintaining similar computational cost.
In addition, a novel network architecture, Balanced HRNet (BHRNet) is proposed to reduce computational costs by reconfiguring the proper number of convolutional blocks on each branch of the network for a balanced distribution of computational costs.
Finally, the DIR module is integrated into the BHRNet architecture for developing the real-time MPPE network, i.e., DIR-BHRNet.
The framework of DIR-BHRNet is compared with the typical MPPE network as shown in \autoref{fig:system-setup}.

We evaluate the novel DIR module and BHRNet architecture on the well-known COCO and CrowdPose datasets.
The experimental results demonstrate that both the DIR module and BHRNet architecture contribute to the real-time MPPE.
In DIR, both the extra depthwise convolution and shortcut connection contribute to higher accuracy.
For the BHRNet architecture, DIR-BHRNet outperforms the network in that DIR modules are integrated within the well-known HRNet architecture (DIR-HRNet).
Furthermore, we compare the DIR-BHRNet with the state-of-the-art MPPE methods, such as EfficientHRNet \cite{neff2020efficienthrnet}, MobileNet V2 \cite{sandler2018mobilenetv2}, and Lightweight OpenPose \cite{cao2019openpose}.
The results show that DIR-BHRNet significantly outperforms both MobileNet V2 and Lightweight OpenPose in terms of both accuracy and computational costs, and EfficientHRNet in terms of accuracy with twice GFLOPS. 
Finally, we install and test the DIR-BHRNet on the current mainstream Android smartphones.
DIR-BHRNet performs more than 10 FPS on these smartphones.
The free-used executable file (Android 10), source code, and video description are publicly available.

The main contributions of this work for facilitating the real-time MPPE development on smartphones are summarized:
\begin{enumerate}
    \item a novel convolutional module, DIR, with an extra depthwise convolution and shortcut connection to enhance the spatial feature for improving accuracy.
    \item a novel network structure, BHRNet, with a balanced distribution of computation by reconfiguring the number of convolutional blocks to reduce computational costs.
    \item the publicly available source code, executable file (Android 10), and implementation on Android smartphones.
\end{enumerate}

\vspace{-5pt}
\section{Related Work}
\label{sec:related}

In this section, we review the related work from real-time HPE and on-device MPPE. 

\vspace{-11pt}
\subsection{Real-time Human Pose Estimation}
\vspace{-3pt}

Although the state-of-the-art HPE approaches \cite{cao2019openpose,cheng2020higherhrnet} have performed advanced accuracy on datasets, they generally cost a lot of computation and hardly run on low-performance computation hardware.
OpenPose \cite{cao2019openpose} and AlphaPose \cite{fang2022alphapose} first proposed real-time bottom-up and top-down MPPE methods, but these networks are run on powerful GPU systems.
Although Osokin \cite{osokin2018real} modified the costly VGG backbone of OpenPose to a lightweight MobileNet-based backbone for real-time MPPE, the network was still run on a powerful Intel CPU mobile device.
Neff et al. \cite{neff2020efficienthrnet} combined model scaling and HRNet into a network for bottom-up MPPE, and run the network on powerful NVIDIA GPU systems.
Yu et al. \cite{yu2021lite} developed a Lite-HRNet by applying the efficient shuffle block of ShuffleNet \cite{zhang2018shufflenet} to HRNet, however the network was run on a super powerful 8 NVIDIA V100 GPU system. 
Newell et al. \cite{newell2017associative} proposed the associative embedding method to predict tag values for each keypoint and group them by connecting the keypoint pair with minimal Euclidean distance between tag values, which did not consider the real-time applications.
Many approaches \cite{xiao2018simple} applied off-the-shelf detectors to estimate the bounding box of each human instance and followed by single-person pose estimation for the cropped bounding box regions, which investigated the MPPE accuracy and conducted on powerful GPU systems. 
Recently, Zhuang et al. \cite{zhuang2024fastervoxelpose} proposed a real-time 3D pose estimation FasterVoxelPose+ by constructing a voxel feature volume and an Encoder-Decoder network.
Although these well-known studies contribute to real-time HPE, these networks take a lot of computational costs and are conducted on powerful GPU systems. 
There is a lack of studies that conducted lightweight networks for real-time MPPE on smartphones.

\vspace{-11pt}
\subsection{On-Device MPPE}
\vspace{-3pt}

The current studies focus on reducing the computational costs of deep neural networks for developing real-time MPPE on mobile devices. 
Many studies manually designed or applied automated machine learning approaches to design advanced lightweight networks from scratch, such as MobileNets \cite{sandler2018mobilenetv2}, ShuffleNets \cite{zhang2018shufflenet}, and EifficientNet \cite{tan2019efficientnet}.
In addition, compression techniques, including compression-compilation \cite{cai2021yolobile}, pruning \cite{he2017channel}, \cite{shen2021towards} and knowledge distillation \cite{chen2020adabert} are used to design lightweight networks based on well-known networks.
However, these networks are conducted on powerful computers rather than smartphones, such as EfficientNet on a server-level Intel Xeon CPU computer. 

Some studies utilized temporal technologies to accelerate HPE.
DeepMon \cite{huynh2017deepmon} and DeepCache \cite{xu2018deepcache} cached internal calculated results in each layer for previous frames and then merged similar parts between frames for acceleration. 
Many other studies leverage the hardware characteristics to accelerate neural network inference.  
$\mu$Layer \cite{kim2019mulayer} proposed channel-wise workload distribution for each layer of neural networks and used processor-friendly quantization on CPUs and GPUs for different data types. 
Recently, Seunghyeon et al. \cite{seo2023mdpose} proposed a single-stage instance-aware pose estimation framework,  MDPose, by modeling the joint distribution of human keypoints with a mixture density model. 
Jiang et al. \cite{jiang2024yolo} proposed a direct regression-based human pose estimation YOLO-Rlepose that leverages Transformer to capture global dependencies for keypoint detection through a multi-head self-attention mechanism.
Tao et al. \cite{jiang2023rtmpose} proposed a top-down real-time MPPE network RTMPose that achieved a state-of-the-art performance with 72.2\% AP on COCO and 70+ FPS on a Snapdragon 865 chip. 
However, the computational cost of top-down methods such as RTMPose and Baidu PaddlePaddle TinyPose could significantly increase over the increasing number of detected persons.

\vspace{-8pt}
\section{Methodology}
\label{sec:method}

In this section, we address the novel lightweight DIR-BHRNet in detail, including the DIR module, the balanced HRNet architecture, and loss function.

\vspace{-11pt}
\subsection{Dense Inverted Residual}
\vspace{-3pt}

We propose a novel Dense Inverted Residual module that 
\begin{figure}[!ht] \centering
    \includegraphics[width=.35\textwidth,trim={360 465 380 335},clip]{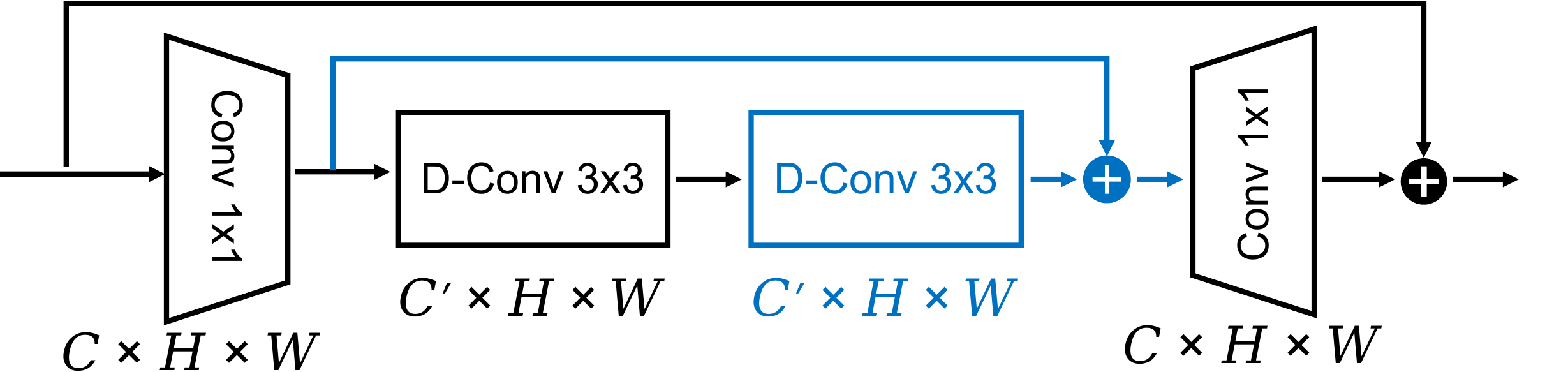} \vspace{-3pt}
    \caption{The DIR architecture. The light blue and black parts represent the added and typical IR, respectively.}
    \label{fig:dir-plus}
\end{figure}
adds an extra depthwise 3x3 convolution (D-Conv $3\times 3$) and a shortcut connection into the well-known Inverted Residual for improving accuracy with a negligible increase of computational cost, as shown in \autoref{fig:dir-plus}.
In this subsection, we first address the basic of Inverted Residual and then the details of the DIR module.

\subsubsection{Inverted Residual} 
Inverted Residual is originally introduced in MobileNet V2 \cite{sandler2018mobilenetv2}.
It consists of two $1\times 1$ standard convolutions, one $3\times 3$ depthwise convolution and a shortcut connection, as the black parts shown in \autoref{fig:dir-plus}.
The standard convolutions are expected to extract features across different channels, yet depthwise convolution is used to extract spatial features.
In Inverted Residual, the depthwise convolution and the standard convolution are expected to extract spatial and channel information, respectively. 
The depthwise convolution and the shortcut connection have shown advanced performance in lightweight deep neural networks \cite{sandler2018mobilenetv2}.

\subsubsection{Dense Inverted Residual (DIR)}
The Dense Inverted Residual is expected to improve the performance of spatial feature extraction while bringing minor extra computational costs.
In general, spatial information could effectively improve the performance of MPPE networks.
In DIR as shown in \autoref{fig:dir-plus}, we therefore add a $3\times 3$ depthwise convolution to enhance the spatial information as an improved IR. 
Furthermore, we add an extra shortcut connection between the two standard convolutions as it prevents gradient confusion with a negligible increase in computational costs.

For computational costs, the number of parameters and computational cost of a standard $3\times 3$ convolution with shape {\small $C$$\times$$H$$\times$$W$} are {\small $C^2$$\times$$3^2$} and {\small $H$$\times$$W$$\times$$C^2$$\times$$3^2$}.
However, for a $3\times 3$ depthwise convolution, the number of parameters and computational cost are {\small $C$$\times$$3^2$} and {\small $H$$\times$$W$$\times$$C$$\times$$3^2$}, respectively.
A $k\times k$ depthwise convolution takes $1/C$ computational cost of a $k$$\times$$k$ standard convolution, and a shortcut connection requires computationally negligible addition.
Therefore, the DIR combining a depthwise convolution and a standard convolution extracts spatial and inter-channel features with merely {\small $1/3^2$$+$$1/C$} computational cost of a standard $3$$\times$$3$ convolution. 

\vspace{-10pt}
\subsection{Balanced HRNet}
\label{sec:method-bhrnet}
\vspace{-4pt}

\subsubsection{DIR-HRNet}
\label{subsec:lite-hrnet}

HRNet is a state-of-the-art backbone network for human pose estimation \cite{sun2019deep}.
In this work, we first design a bottom-up MPPE network by combining DIR modules with HRNet architecture, i.e., DIR-HRNet.
We develop the DIR-HRNet by proposing the novel DIR module and referring to the original HRNet architecture.
HRNet generally uses a high-resolution stream in the first stage. 
The stride-2 convolutions are used as downsampling operations to create lower-resolution branches in the later stages. 
In the end stage, downsampling and upsampling operations are used to exchange and fuse the information.
All stages employ residual modules as the basic convolution block.

Followed by the final stage, we use the state-of-the-art head module of Higher-HRNet \cite{cheng2020higherhrnet} which consists of one deconvolution layer and three residual modules to produce heatmaps and tagmaps.
We chose the associative embedding \cite{newell2017associative} to group predicted keypoints into human instances, in which each keypoint is identified by taking the local peak of heatmaps. 
Finally, the grouping algorithm connects the keypoint pairs with minimal Euclidean distance of tagmap values.
In DIR-HRNet, we implement a lightweight DIR-HRNet by substituting the residual modules in the HRNet backbone with the proposed DIR modules.
In our preliminary experiments, we note that the computation distribution of an n-stage lightweight HRNet is rather biased over feature maps with different resolutions, as shown in \autoref{tab:computaion-dist}. 
The biased computation distribution could have computational redundancy and is improved for efficient MPPE in our lightweight network.

\begin{table}[!ht] \centering \footnotesize
    \renewcommand{\arraystretch}{0.8} \setlength\tabcolsep{2pt}
    \begin{tabular}{c c c >{\columncolor{gray!30}}c c >{\columncolor{gray!30}}c c >{\columncolor{gray!30}}c} \toprule
        \multirow{2}{*}{Resolution} & \multirow{2}{*}{Layer} & \multicolumn{2}{c}{\#Channel} & \multicolumn{2}{c}{\#Conv Blocks} & \multicolumn{2}{c}{Cost} \\ \cline{3-8}
        & & HRNet & BHRNet & HRNet & BHRNet & HRNet & BHRNet \\ \midrule
        \multirow{4}{*}{$\frac{1}{4}\times \frac{1}{4}$}  & stage 1 & 32 & 32 & 2 & 1 & \multirow{4}{*}{40\%} & \cellcolor{gray!30} \\ 
        & stage 2  & 32 & 32 & 2 & 1 & & \cellcolor{gray!30} \\ 
        & stage 3  & 32 & 32 & 2 & 1 & & \cellcolor{gray!30} \\ 
        & stage 4  & 32 & 128 & 2 & 1 & & \cellcolor{gray!30} \multirow{-4}{*}{20\%}\\  \midrule
        \multirow{3}{*}{$\frac{1}{8}\times \frac{1}{8}$} & stage 2 & 64 & 64 & 2 & 2 &\multirow{3}{*}{ 30\%} & \cellcolor{gray!30} \\ 
        & stage 3  & 64 & 64 & 2 & 2 & & \cellcolor{gray!30} \\ 
        & stage 4  & 64 & 128 & 2 & 2 & & \cellcolor{gray!30} \multirow{-3}{*}{30\%}\\ \midrule
        \multirow{2}{*}{$\frac{1}{16}\times \frac{1}{16}$} & stage 3 & 128 & 128 & 2 & 3 & \multirow{2}{*}{20\%} & \cellcolor{gray!30} \\ 
        & stage 4  & 128 & 128 & 2 & 3 & & \cellcolor{gray!30} \multirow{-2}{*}{20\%} \\ \midrule
        $\frac{1}{32}\times \frac{1}{32}$ & stage 4 & 256 & 128 & 2 & 4 & 10\% & 20\% \\ \bottomrule
    \end{tabular} \vspace{-3pt}
    \caption{Computation costs allocation on feature maps with different resolutions in HRNet and BHRNet.}
    \label{tab:computaion-dist}
\end{table} \vspace{-5pt}

\subsubsection{DIR-BHRNet}
\begin{figure}[!ht] \centering \small
    \includegraphics[width=0.48\textwidth,trim={5 5 10 5},clip]{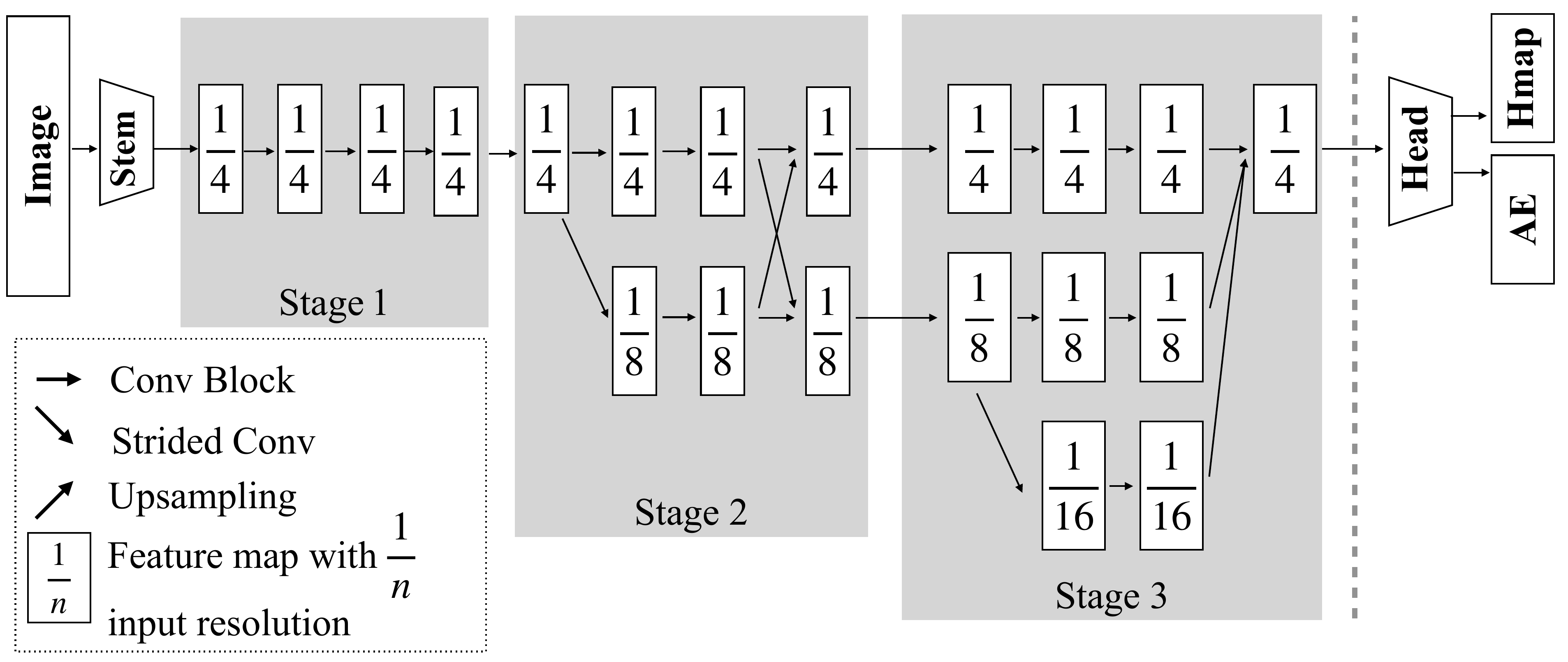} \vspace{-3pt}
    \caption{A diagram of DIR-BHRNet built by novel DIR modules and BHRNet architecture. $\frac{1}{n}$ represents the DIR modules.}
    \label{fig:our-net}
\end{figure}

As shown in \autoref{tab:computaion-dist}, the HRNet backbone performs a biased computation distribution, which specifically takes $40\%$ of total computational resources on the feature maps with the largest resolution.
In contrast, the feature map with the smallest feature map merely costs $10\%$ computational resources. 
In this work, we propose a novel balanced HRNet (BHRNet) which reconfigures the number of convolutional blocks for a balanced distribution of computational costs and reduces computational costs.
We integrate the DIR module and BHRNet architecture to build the lightweight network, DIR-BHRNet.
\autoref{fig:our-net} shows the architecture of a DIR-BHRNet.

Instead of setting an equal number of convolutional blocks 
for each branch at every stage of an HRNet-structured network, we reconfigure the number of convolutional blocks so the computational cost on each branch is approximately equal.
In this work, we design a 4-stage Balanced HRNet architecture as the original HRNet, which performs a balanced computational cost on each branch, as shown in \autoref{tab:computaion-dist}.

We use a head module to produce heatmaps and tagmaps followed by the final stage of BHRNet. 
In contrast to the DIR-HRNet, we use two different head modules to fully investigate the efficiency of DIR-BHRNet.
The first one is the well-known head module of Higher-HRNet \cite{cheng2020higherhrnet}, which produces higher accuracy but brings extra computational costs.
The second one consists of one single convolution layer. 

\vspace{-9pt}
\subsection{Loss Function}
\vspace{-3pt}

In this work, our bottom-up methods identify all joint keypoints and group them into personal instances.
For each personal instance in multi-person pose estimation, the network generates {\small $K$$\times$$H$$\times$$W$} keypoint heatmaps, where {\small $K$} is the defined number of keypoints for each person (e.g., {\small $K$$=$$17$} in the COCO dataset).
Therefore, the heatmap value of $K$ keypoint for $n$ person can be calculated by:
\begin{equation} \small
    \mathcal{H}_{k,n}(p) = e^{\frac{\|p-p_{k,n}^{\prime}\|_{2}^2}{\sigma^2}}, ~ \forall k=1,2,\ldots,K, ~ \forall n=1,2,\ldots,\mathcal{N} \\
    \label{eq:multi-personHeatmap}
\end{equation}
where $p$ represents the position $(x,y)$ in the heatmap, $p^{\prime}_{k,n}$ represents the position $(x^{\prime},y^{\prime})$ of $k_{th}$ keypoint of $n_{th}$ person in the heatmap, $\sigma$ is a constant that is 2 in this work as HRNet. The position with a maximum value in the heatmap is the joint keypoint.
Therefore, the heatmap value of the keypoints can be calculated by:
\begin{equation} \small
    \mathcal{H}^{\prime}_k(p) = \max_{n} \mathcal{H}_{k,n}(p), \forall n=1,2,\ldots,\mathcal{N}
     \label{eq2}
\end{equation}

First, we calculate the loss function for predicting joint keypoints by considering the mean squared error between the predicted keypoint heatmaps {\small $\mathcal{H}_k \in \mathbb{R}^{H \times W}$} and the ground truth {\small $\mathcal{H}_k^* \in \mathbb{R}^{H \times W}$}.
Therefore, the loss function {\small ($\mathcal{L}_{\mathcal{H}}$)} for predicting the keypoint heatmaps can be calculated by:
\begin{equation} \footnotesize
    \mathcal{L}_{\mathcal{H}} = \frac{1}{K\cdot H\cdot W} \sum_{k=1}^K \sum_{p\in \mathbb{Z}^{H\times W}}(\mathcal{H}^{\prime}_k (p) - \mathcal{H}_k^*(p))^2
\end{equation}

Second, we consider the tag error in the loss function to minimize the tag error of the keypoints for a personal instance and maximize the labeling error of the keypoints from different personal instances.
For the predicted tags $\mathcal{H}_k \in \mathbb{R}^{H\times W}$, the loss function $\mathcal{L}_T$ with tag error can be calculated by:
\begin{equation} \footnotesize
    \mathcal{L}_{\mathcal{T}} = \frac{1}{N} \sum_{n} \sum_{k} ( \mathcal{T}_k (p_{n,k}) - \bar{\mathcal{T}}_n)^2    + \frac{1}{N^2} \sum_{n}\sum_{n'\neq n} e^{ - ( \bar{\mathcal{T}}_n - \bar{\mathcal{T}}_{n'} )^2/2\sigma^2}    
\end{equation}
where $p_{k,n}$ is the position of $k_{th}$ keypoints for $n_{th} (n\in \mathbb{Z}^N)$ person, $\bar{\mathcal{T}_n}=\frac{1}{K} \sum_{k=1}^K \mathcal{T}_{p_{n,k}}$ represents the mean tag value of $n_{th}$ person.
The final loss function intergrates $\mathcal{L}_{\mathcal{H}}$ and $\mathcal{L}_T$ by the proper weight as:
\begin{equation} \centering \small
    \mathcal{L} = \alpha \mathcal{L}_{\mathcal{H}} + \beta \mathcal{L}_{\mathcal{T}} 
    \label{eq:overall-loss}
\end{equation}
where $\alpha$ and $\beta$ are the weights that should be tuned during the training process since $\mathcal{L}_{\mathcal{H}}$ and $\mathcal{L}_T$ perform different scales.
In this work, $\alpha$ and $\beta$ are set as 0.99 and 0.01 respectively after the parameter tuning in our preliminary experiments, since $\mathcal{L}_{\mathcal{H}}$ and $\mathcal{L}_T$ perform the value at significant different scales.

\vspace{-5pt}
\section{Experiments}
\label{sec:expe}
\vspace{-3pt}

In this section, we address the details of the experiments for testing our methods and the state-of-the-art methods. 
We conduct extensive comparison experiments and ablation studies on the well-known multi-person pose estimation datasets: COCO \cite{lin2014microsoft} and CrowdPose \cite{li2019crowdpose}. 
To validate the performance on crowded scenes, we evaluate the methods on the CrowdPose dataset which contains more crowded scenes than COCO.
Specifically, we investigate the effect of the lightweight DIR module and the architecture of HRNet and balanced HRNet, i.e., DIR-HRNet and DIR-BHRNet.
Finally, to demonstrate the performance of our lightweight networks on low-performance computing hardware, we implement the DIR-HRNet and DIR-BHRNet on Android smartphones via Android Studio. 

\vspace{-8pt}
\subsection{Datasets and Evaluation Metrics}
\label{subsec:dataset}

\subsubsection{Datasets}

In this work, we evaluate the methods on the well-known datasets: COCO \cite{lin2014microsoft} and CrowdPose \cite{li2019crowdpose}.
The COCO dataset consists of over 200k images and 250k annotated person instances, which are divided into three sets, \textit{train2017}, \textit{val2017}, and \textit{test-dev2017} that contain 57k, 5k and 20k images respectively.
We train the networks on the COCO \textit{train2017} set and evaluate the trained networks on the COCO \textit{val2017} and COCO \textit{test-dev2017}.
The CrowdPose dataset contains over 20k images and 80k annotated person instances. 
The dataset is split into the training, validation, and testing sets with a ratio of 5:1:4. 
Note that our methods are evaluated and compared with the state-of-the-art studies on COCO and CrowdPose datasets.

Furthermore, we apply data augmentation of random rotation ($[-30^\circ,30^\circ]$), random scale $([0.75,1.5])$, and random translation $([-40,40])$ to enrich the training datasets \cite{newell2017associative,cheng2020higherhrnet}. 
In addition, we resize the image to various sizes (e.g., $256\times 256$, $384\times 384$, $512\times 512$) for both training and testing. 

\subsubsection{Evaluation Metrics}
As the common metrics of the COCO and CrowdPose, we use the Object Keypoint Similarity ($\mathcal{OKS}$)-based mean average precision (mAP) as our primary metric on both datasets.
For more details about $\mathcal{OKS}$, please refer to the studies of COCO \cite{lin2014microsoft} and CrowdPose \cite{li2019crowdpose}.
Furthermore, we evaluate the lightweight networks in terms of GFLOPS and FPS on the datasets and smartphones.

\vspace{-11pt}
\subsection{Experiments on Datasets}
\vspace{-5pt}
In this subsection, we present detailed experimental setups for DIR-HRNet and DIR-BHNRet.
The neural networks are trained by using the Adam optimizer, in which the learning rate is set to $1.5e-3$ at the beginning and drops to $1e-4$ and $1e-5$ at the 200th epoch and 250th epoch, respectively.  
The whole training process is conducted on PyTorch.
The images are resized to the same size as the input in the training setup. 
We use flip testing in all the experiments \cite{cheng2020higherhrnet}. 

\subsubsection{DIR-HRNet}

We investigate the effect of modules in DIR by testing various DIR setups.
Specifically, we investigate the effect of each component of the DIR module by comparing the performance of four basic convolutional blocks: 1) IR, 2) the combination of IR and an extra depthwise convolution, 3) the combination of IR and an extra skip connection, 4) the combination of IR, an extra depthwise convolution and an extra skip connection.
In addition, we conduct experiments to tune input size.
We set the network width to 32 and the number of stages to 4 according to the original HRNet \cite{sun2019deep}.
Finally, we choose the best performance setup to compare with the state-of-the-art methods on the COCO and CrowdPose datasets. 

\subsubsection{DIR-BHRNet}
We develop a novel DIR-BHRNet by replacing the structure of HRNet in DIR-HRNet with the BHRNet structure. 
We use the same hyperparameter setup of DIR-HRNet to DIR-BHRNet (e.g., network width, the number of network stages).
To investigate the effectiveness of DIR-BHRNet, we apply two representative modules in the head module of DIR-BHRNet, the well-known HigherHRNet head module \cite{cheng2020higherhrnet} and a head module of a single 3x3 convolutional layer due to their different performance in terms of accuracy and computational costs. 
Finally, we compare the best DIR-BHRNet with DIR-HRNet and the state-of-the-art methods on the COCO and CrowdPose datasets.

\vspace{-8pt}
\subsection{Experiments on Smartphones}
\vspace{-3pt}

To demonstrate the lightweight DIR-BHRNet on smartphones for MPPE, 
we develop the Android executable file on the integrated development environment of Android Studio and visualize the processed output/poses through Android AnativeWindow API to the user interface for evaluations.
Specifically, we convert the trained DIR-BHRNet into ONNX format, which is readable for various machine learning frameworks (e.g., PyTorch, TensorFlow and Keras).
In this work, we implement the DIR-BHRNet based on the NCNN framework \cite{ncnn}. 
Currently, NCNN is the fastest, cross-platform, third-party dependency-free, and open-source framework on mobile phone CPUs.
NCNN is a high-performance framework that is extremely optimized for mobile platforms.
Furthermore, NCNN is cross-platform in C++ without any third-party dependencies and runs faster than the current well-known open-source frameworks on mobile platforms.
Finally, NCNN covers the most well-known CNN networks such as the lightweight CNNs of MobileNet and ShuffleNet.

\begin{figure}[!ht] \centering \footnotesize
    \begin {tikzpicture}
        \node[draw,align = center,minimum height = 23pt] at (-4.8, 0)  (a) {Trained \\ DIR-BHRNet};
        \node[draw,align = center,minimum height = 23pt] at (-2.4, 0)   (b) {ONNX};
        \node[draw,align = center, minimum height = 23pt,text depth = 18pt] at (0.0, 0)   (c) {Android Studio};
        \node[draw,align = center, minimum height = 12pt] at (-0.5, -0.16)  (i) {C++};
        \node[draw,align = center, minimum height = 12pt] at (0.5, -0.16)  (j) {JAVA};
        \node[draw,align = center, minimum height = 23pt] at (2.2, 0)   (d) {APK \\ (smartphones) };
        \node[draw,align = center, minimum height = 20pt] at (0.0, 1.2)   (e) {CPU Profiler};
        \node[draw,align = center, minimum height = 20pt,fill=gray!20] at (-2.1, 1.2)   (f) {Memory usage};
        \node[draw,align = center, minimum height = 20pt,fill=gray!20] at (2.1, 1.2)   (g) {Latency};
        \draw[->,semithick] (a) node[above,xshift=1.4cm] {Pytorch} -- (b);
        \draw[->,semithick] (b) node[above,xshift=1.0cm] {NCNN} -- (c);
        \draw[->,semithick] (c) node[above,xshift=1.55cm] {} -- (d);
        \path (c.east) -- (c.north east) coordinate[pos=0.4] (c1);
        \path (d.west) -- (d.north west) coordinate[pos=0.5] (d1);
        \draw[->,semithick] (d1) node[above,xshift=1.55cm] {} -- (c1);
        \draw[->,semithick] (c) node[above,xshift=1cm] {} -- (e);
        \draw[->,semithick] (e) node[above,xshift=1cm] {} -- (f);
        \draw[->,semithick] (e) node[above,xshift=1cm] {} -- (g);
    \end {tikzpicture} \vspace{-13pt}
\caption{The implementation and evaluation framework of our MPPE networks on smartphones.} \label{fig:expeirments} 
\end{figure}
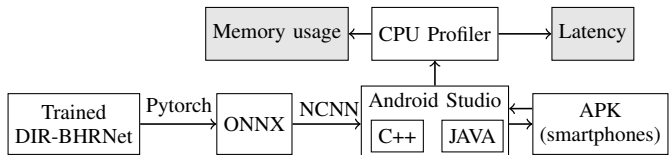

We implement MPPE program on Android smartphones and collect the computational cost of the MPPE program via the CPU Profiler of Android Studio that connects the Android smartphones.
We record the latency and used memory which are the two common factors to evaluate practical MPPE applications. 
The experimental framework of implementation and evaluation on smartphones are shown in \autoref{fig:expeirments}.
We test the networks on the representative Android smartphones of Xiaomi 10, Xiaomi 12, and Redmi K40 with the CPUs (Memory) of Snapdragon 865 (8GB), Snapdragon 870 (8GB), and Snapdragon 8 (12GB) respectively.

\vspace{-11pt}
\section{Results and Discussion}
\label{sec:results}
\vspace{-3pt}

In this section, we present the experimental results on the datasets and Android smartphones.
The two proposed networks, DIR-based HRNet and DIR-BHRNet, are tested on the COCO and CrowdPose datasets.
We investigate ablation studies on datasets and compare the results with state-of-the-art methods. 
Finally, we investigate the real-time performance and computational cost of our lightweight networks on Android smartphones.

\vspace{-11pt}
\subsection{Results on Datasets}
\vspace{-3pt}

\subsubsection{DIR-HRNet}

We validate the motivation for adding the depthwise convolutions in DIR-HRNet: adding more spatial extraction for higher accuracy.
We observe the performance of employing a different number of depthwise $3\times3$ convolutions in a DIR module, the results on the COCO and CrowdPose datasets as shown in \autoref{tab:num_blocks}.
The DIR-HRNet with two depthwise convolutions performs the best mAP of 40.3 and 38.6 on COCO and CrowdPose datasets respectively.
The results show that the DIR-HRNet with two depthwise convolutions outperforms the DIR-HRNets with the configuration of the other number of depthwise convolutions.

Furthermore, we investigate the effect of each component in DIR via ablation studies.
Specifically, we conduct the experiments and compare the result of different lightweight networks based on four basic convolutional blocks as shown in \autoref{tab:ablation-DIR-coco}.
The results show that both an extra depthwise convolution and an extra shortcut connection in DIR contribute to significant mAP improvements.
On the COCO dataset, while the components of an extra depthwise convolution and shortcut connection improve the mAP value by 1.0 and 0.4 respectively, the proposed DIR with the combination of depthwise convolution and shortcut connection significantly improves the mAP value by 3.2. 
On the CrowdPose dataset, while the single extra component hardly improves the performance, our DIR with the combination of depthwise convolution and shortcut connection performs an improved mAP value of 1.6.
The four convolutional blocks perform similar GFLOPS values, which demonstrates the extra depthwise convolution and shortcut connection cost negligible computational.
The combination of IR and an extra depthwise convolution performs the same computational cost of 2.36 GFLOPS with IR.
The DIR module with an extra depthwise convolution and an extra shortcut connection performs 2.44 GFLOPS which increases by $3\%$ over the IR module.

\begin{table}[!ht] \centering \small \renewcommand{\arraystretch}{0.8}
    \begin{tabular}{l|c|c|c|c} \toprule
        Dataset & \#D-Convs & \#Params & GFLOPs & mAP \\ \midrule
        \multirow{4}{*}{COCO} & 1 & 2.43 M & 2.36 & 37.5 \\
        & \cellcolor{gray!30}2 & \cellcolor{gray!30}2.51 M & \cellcolor{gray!30}2.44 & \cellcolor{gray!30}\textbf{40.3} \\
        & 3 & 2.58 M & 2.52 & 38.0 \\
        & 4 & 2.65 M & 2.60 & 32.7 \\ \midrule
        \multirow{4}{*}{CrowdPose} & 1 & 2.43 M & 2.36 & 37.1 \\
        & \cellcolor{gray!30}2 & \cellcolor{gray!30}2.51 M & \cellcolor{gray!30}2.44 & \cellcolor{gray!30}\textbf{38.6} \\
        & 3 & 2.58 M & 2.52 & 38.0 \\
        & 4 & 2.65 M & 2.60 & 32.5 \\ \bottomrule
    \end{tabular} \vspace{-3pt}
    \caption{The results of DIR-HRNet on the COCO and CrowdPose datasets with different numbers of depthwise 3x3 convolutions in a basic convolutional block.}
    \label{tab:num_blocks}
\end{table} \vspace{-5pt}

\begin{table*}[!ht] \centering \small
    \renewcommand{\arraystretch}{0.8}
    \setlength\tabcolsep{5pt} 
    \begin{tabular}{c| l |c | c | c| c| c| c| c}  \toprule
         Dataset & Conv Block & \#Params & GFLOPS &mAP& $\text{AP}^{50}$ & $\text{AP}^{75}$ & $\text{AP}^{M}$ & $\text{AP}^{L}$\\ \midrule
         \multirow{5}{*}{COCO} & IR + Squeeze-excite connection & 2.73M & 2.46 & 40.1 & 68.1 & 39.4 & 29.1 & 55.2 \\
         & IR & 2.43M & 2.36 & 37.1 & 64.7 & 36.7 & 27.2 & 51.1\\
         & IR + Extra depthwise Convolution & 2.51M & 2.44 & 38.1 & 65.8 & 37.7 & 27.7 & 53.1\\
         & IR + Extra shortcut connection & 2.43M & 2.36 & 37.5 & 65.4 & 37.4 & 27.2 & 52.4 \\
         & \cellcolor{gray!30}IR + Extra depthwise Convolution \& shortcut connection & \cellcolor{gray!30}2.51M & \cellcolor{gray!30}2.44 & \cellcolor{gray!30}\textbf{40.3} &\cellcolor{gray!30}\textbf{68.4} & \cellcolor{gray!30}\textbf{40.1} & \cellcolor{gray!30}\textbf{29.6} & \cellcolor{gray!30}\textbf{56.1} \\ \midrule
         \multirow{4}{*}{CrowdPose} & IR & 2.43M & 2.36 & 37.0 & 66.3 & 34.7 & 36.2 & 49.5 \\
         & IR + Extra depthwise Convolution & 2.51M & 2.44 & 36.9 & 66.6 & 34.4 & 36.3 & 49.0 \\
         & IR + Extra shortcut connection & 2.43M & 2.36 & 37.1 & 66.7 & 35.1 & 36.6 & 49.1 \\
         & \cellcolor{gray!30}IR + Extra depthwise Convolution \& shortcut connection & \cellcolor{gray!30}2.51M & \cellcolor{gray!30}2.44 & \cellcolor{gray!30}\textbf{38.6} & \cellcolor{gray!30}\textbf{68.3} & \cellcolor{gray!30}\textbf{36.9} & \cellcolor{gray!30}\textbf{38.3} & \cellcolor{gray!30}\textbf{50.1} \\ \bottomrule
    \end{tabular} \vspace{-3pt}
    \caption{The results of ablation studies on the components in our novel DIR module.}
    \label{tab:ablation-DIR-coco}
\end{table*} \vspace*{-7pt}

Furthermore, we investigate the effects of different input size values in DIR-HRNet.
The results show that the DIR-HRNet with the input size of $384$$\times$$384$ performs the best mAP on both COCO and CrowdPose datasets, as shown in \autoref{tab:ablation-input-size}.
Generally, DIR-HRNet performs higher mAP and GFLOPs on large inputs than small ones.
Particularly, DIR-HRNet performs a slightly lower mAP value with the input size $512$$\times$$512$ than the one with input size $384$$\times$$384$.

Finally, we compare the performance of DIR-HRNets and the state-of-the-art methods on COCO 2017val and CrowdPose val datasets.
The comparative results are shown in \autoref{tab:BHRNet-comparison}. 
As in the same notation, we use DIR-HRNet-\(k\) to denote DIR-HRNet networks with width \(k\).
Compared to MobileNet V2 and Lightweight OpenPose, DIR-HRNet-32 performs better performance with $125.6\%$ mAP and $111.7\%$ of their average accuracy and $61\%$ (5.49/9.0) of their GFLOPS.

\subsubsection{Balanced HRNet} 
We observe the DIR-HRNet and DIR-BHRNet training process on the COCO and CrowdPose datasets.
Both DIR-BHRNet and DIR-HRNet converge to a stable loss value for fair comparison. 
DIR-BHRNet converges to a lower loss than DIR-HRNet within 300 training epochs, which generally indicates the performance of higher accuracy.

Furthermore, we compare the performance of the trained DIR-HRNet and DIR-BHRNet with different input sizes from the aspects of GFLOPS and mAP, as shown in \autoref{tab:crowdpose-comparison}.
The DIR-BHRNet significantly outperforms DIR-HRNet in terms of accuracy with approximately $15\%$ extra computational costs which come from the rounding up operation in determining the number of basic convolutional blocks.
\begin{table}[!ht] \centering \small \renewcommand{\arraystretch}{0.8}
    \begin{tabular}{c | c| c | c | c}  \toprule
        Dataset & Input Size & \#Params & GFLOPS & mAP \\  \midrule
        \multirow{4}{*}{COCO} & $128\times 128$ &2.51M& 0.61 & 14.9 \\ 
        & $256\times 256$ &2.51M& 2.44 &40.3 \\
        & \cellcolor{gray!30}$384\times 384$ & \cellcolor{gray!30}2.51M & \cellcolor{gray!30}5.49 & \cellcolor{gray!30} \textbf{47.8} \\ 
        & $512\times 512$ &2.51M& 9.77 & 47.4 \\
        \midrule
        \multirow{4}{*}{CrowdPose} & $128\times 128$ &2.51M& 0.61 & 18.2 \\ 
        & $256\times 256$ &2.51M& 2.44 &40.3 \\ 
        & \cellcolor{gray!30}$384\times 384$ & \cellcolor{gray!30}2.51M & \cellcolor{gray!30}5.49 & \cellcolor{gray!30} \textbf{45.2} \\ 
        & $512\times 512$ &2.51M& 9.77 & 39.3 \\ 
    \bottomrule
    \end{tabular} \vspace{-3pt}
    \caption{The experimental results of DIR-HRNet with a width of 32 on COCO 2017val and CrowdPose val.}
    \label{tab:ablation-input-size}
\end{table} \vspace{-5pt}
We investigate the efficiency of DIR-BHRNet by reducing the network width to observe the GFLOPs and accuracy of DIR-BHRNet.
Specifically, we reduce the network width of DIR-BHRNet from the default value of 32 to 25 (denoted as DIR-BHRNet-25).
As the results in \autoref{tab:BHRNet-comparison} and \autoref{tab:crowdpose-comparison}, DIR-HRNet and DIR-BHRNet perform the trade-off between the values of GFLOPs and FPS when the input size is $384\times384$.
\begin{table*}[!b] \centering \small
    \renewcommand{\arraystretch}{0.8} \setlength\tabcolsep{1.5pt} 
    \begin{tabular}{l l|l|l|c|c|c|c|c}  \toprule
    Dataset & Networks & Backbone & Grouping Algorithms & Input Size & \#Params & \footnotesize{GFLOPS} & Latency & mAP \\  \midrule
    \multirow{12}{*}{COCO} & E2Pose \cite{tobeta2022e2pose} & ResNet & E2Pose\cite{tobeta2022e2pose} & 320x320 & 23.9M & - & 54.1ms$^\ddagger$ & 43.9 \\
    & YOLObile \cite{cai2021yolobile} & EfficientNet & Part Affinity Fields \cite{cao2019openpose} & 384x384 & 1.68M & 4.7 & 67.2ms & 41.6 \\
    & \multirow{2}{*}{EfficientHRNet \cite{neff2020efficienthrnet}}
     & EfficientHRNet & Associative Embedding \cite{newell2017associative}& 384x384 & 3.7M & 2.1 & 30.0ms & 35.7 \\
     & & EfficientHRNet & Associative Embedding & 416x416 & 6.9M & 4.2 & 60.9ms & 44.8 \\
    & MobileNet V2 \cite{sandler2018mobilenetv2}& MobileNet V2 & Associative Embedding & 512x512 & 9.6M & 8.6 & 121.1ms & 38.0 \\
    & \footnotesize{Lightweight OpenPose} \cite{osokin2018real} & MobileNet V1 & Part Affinity Fields \cite{cao2019openpose}& 368x368 & 4.1M & 9.0 & 38.5ms$^\dagger$ & 42.8 \\ \cline{2-9}
    & DIR-HRNet-24 (ours) & DIR-HRNet(ours) & Associative Embedding & 256x256 & 1.64M & 2.44 & 34.9ms & 36.2 \\
    & DIR-HRNet-48 (ours) & DIR-HRNet & Associative Embedding & 256x256 & 5.51M & 3.68 & 52.6ms & 44.9 \\
    & \cellcolor{gray!30}DIR-HRNet-32 (ours) & \cellcolor{gray!30}DIR-HRNet  & \cellcolor{gray!30}Associative Embedding & \cellcolor{gray!30}384x384 & \cellcolor{gray!30}2.51M & \cellcolor{gray!30}\textbf{5.49}  & \cellcolor{gray!30}79.6ms & \cellcolor{gray!30}\textbf{47.8} \\ \cline{2-9}
    & DIR-BHRNet$^*$-25 (ours) & DIR-BHRNet & Associative Embedding & 384x384 & 3.4 M & 1.91 & 27.3ms & 41.2 \\
    & DIR-BHRNet$^*$-32 (ours) & DIR-BHRNet & Associative Embedding & 384x384 & 5.9 M & 3.23 & 46.2ms & 45.2 \\
    & DIR-BHRNet-25 (ours) & DIR-BHRNet & Associative Embedding & 384x384 & 4.7M & 5.10 & 72.9ms & 49.5 \\
    & \cellcolor{gray!30}DIR-BHRNet-32 (ours) & \cellcolor{gray!30}DIR-BHRNet & \cellcolor{gray!30}Associative Embedding & \cellcolor{gray!30}384x384 & \cellcolor{gray!30}6.03M & \cellcolor{gray!30}\textbf{6.32} & \cellcolor{gray!30}89.0ms & \cellcolor{gray!30}\textbf{50.5 ($5.6\%\uparrow$)} \\ \midrule
    \multirow{6}{*}{CrowdPose} & E2Pose \cite{tobeta2022e2pose} & ResNet & E2Pose \cite{tobeta2022e2pose} & 320x320 & 23.9M & - & 54.1ms$^\ddagger$ & 43.9 \\
    & EfficientHRNet \cite{neff2020efficienthrnet} & EfficientHRNet & Associative Embedding & 416x416 & 6.9M & 4.2 & 60.9ms & 46.1 \\ \cline{2-9}
    & DIR-HRNet-24 (ours) & DIR-HRNet(ours) & Associative Embedding & 256x256 & 1.64M & 2.44 & 34.9ms & 40.3 \\
    & \cellcolor{gray!30}DIR-HRNet-32 (ours) & \cellcolor{gray!30}DIR-HRNet & \cellcolor{gray!30}Associative Embedding & \cellcolor{gray!30}384x384 & \cellcolor{gray!30}2.51M & \cellcolor{gray!30}\textbf{5.49} & \cellcolor{gray!30}79.6ms & \cellcolor{gray!30}\textbf{45.2} \\ \cline{2-9}
    & DIR-BHRNet-25 (ours) & DIR-BHRNet & Associative Embedding & 384x384 & 4.7M & 5.10 & 72.9ms & 50.4 \\
    & \cellcolor{gray!30}DIR-BHRNet-32 (ours) & \cellcolor{gray!30}DIR-BHRNet & \cellcolor{gray!30}Associative Embedding & \cellcolor{gray!30}384x384 & \cellcolor{gray!30}6.03M & \cellcolor{gray!30}\textbf{6.32} & \cellcolor{gray!30}89.0ms & \cellcolor{gray!30}\textbf{52.4(15.9\%)} \\ \bottomrule
    \end{tabular} \vspace{-3pt}
    \caption{The performance of the proposed and the state-of-the-art networks on the COCO 2017val and CrowdPose. $^*$ indicates the networks with a head module of a single 3x3 convolutional layer. The latency values are mean on three smartphones. $\dagger$ was tested on the powerful computer system with CPU i7-6850K. $\ddagger$ was tested on the GPU system Jetson AGX Xavier.}
    \label{tab:BHRNet-comparison}
\end{table*}
Here, we test DIR-BHRNet-25 with an input size of $384\times384$ for observing a comparison, as the results shown in \autoref{tab:crowdpose-comparison}.
Although DIR-BHRNet-25 performs with lower accuracy than DIR-BHRNet-32, outperforms DIR-HRNet in terms of both accuracy and computational costs.

\begin{table}[!ht] \centering \small 
    \setlength\tabcolsep{2pt} \renewcommand{\arraystretch}{0.8}
    \begin{tabular}{l l | c | c | l} \toprule
        Dataset & Model & Input Size & \footnotesize{GFLOPS} & mAP \\ \midrule
        \multirow{7}{*}{\footnotesize{CrowdPose}} & DIR-HRNet & 256x256 & 2.44  & 40.3 \\
        & DIR-BHRNet & 256x256 & 2.81 & 44.0 ($9.2\%\uparrow$) \\  \cmidrule(r){2-5}
        & DIR-HRNet & 384x384 & 5.49 & 45.2 \\
        & \cellcolor{gray!30}DIR-BHRNet & \cellcolor{gray!30}384x384 & \cellcolor{gray!30}6.32 & \cellcolor{gray!30}\textbf{52.4 ($15.9\%\uparrow$)} \\  \cmidrule(r){2-5}
        & DIR-BHRNet-25 & 384x384 & 5.10 & 50.4 ($11.5\%\uparrow$) \\ \cmidrule(r){2-5}
        & DIR-HRNet & 512x512 &  9.77 & 39.3 \\
        & DIR-BHRNet & 512x512 & 11.23 & \cellcolor{gray!30}\textbf{52.8 ($34.4\%\uparrow$)} \\  \midrule
        \multirow{7}{*}{COCO} & DIR-HRNet & 256x256 & 2.44  & 36.2 \\
        & DIR-BHRNet & 256x256 & 2.81 & 39.2 ($8.3\%\uparrow$) \\  \cmidrule(r){2-5}
        & DIR-HRNet & 384x384 & 5.49 & 47.8 \\
        & \cellcolor{gray!30}DIR-BHRNet & \cellcolor{gray!30}384x384 & \cellcolor{gray!30}6.32 & \cellcolor{gray!30}\textbf{50.5 ($5.6\%\uparrow$)} \\  \cmidrule(r){2-5}
        & DIR-BHRNet-25 & 384x384 & 5.10 & 49.5 ($3.6\%\uparrow$) \\ \cmidrule(r){2-5}
        & DIR-HRNet & 512x512 &  9.77 & 47.3 \\
        & DIR-BHRNet & 512x512 & 11.23 & \cellcolor{gray!30}\textbf{53.2 ($12.5\%\uparrow$)} \\ \bottomrule
    \end{tabular} \vspace{-3pt}
    \caption{The comparison results (ablation studies) of DIR-HRNet and DIR-BHRNet on COCO and CrowdPose datasets.}
    \label{tab:crowdpose-comparison}
\end{table} \vspace{-5pt}

Finally, we compare the performance of DIR-BHRNet with state-of-the-art methods on COCO and CrowdPose datasets, as shown in \autoref{tab:BHRNet-comparison}.
The results show that DIR-BHRNet with the head module of a single 3x3 convolutional layer (tagged by *) outperforms the EfficientHRNet models with GFLOPs <= 4.2.
For example, considering EfficientHRNet with 2.1 GFLOPs, the DIR-BHRNet$^*$-25 with 1.91 GFLOPS perform a $9\%$ improvement in terms of GFLOPS and a $15.4\%$ improvement in terms of mAP.
Compared to EfficientHRNet with 4.2 GFLOPs, DIR-BHRNet$^*$-32 performs an approximately $23.1\%$ reduction in GFLOPs (3.23) and a 0.9\% improvement in terms of mAP.
DIR-BHRNet-25 performs 49.5 mAP which is much higher than the state-of-the-art methods, EfficientHRNet, MobileNet, Lightweight OpenPose, and performs 5.10 GFLOPS which is much lower than the MobileNet and Lightweight OpenPose.
Furthermore, DIR-BHRNet-32 performs a higher mAP of 50.5 with 6.32 GFLOPS. 
We observed that DIR-HRNet-32 performs similarly to DIR-BHRNet-32 in terms of GFLOPS (5.49 vs 6.32) and average latency (79.6ms vs 89.0ms, i.e., > 10 FPS). 
DIR-BHRNet-32 outperforms DIR-HRNet-32 with an improvement of 5.6\% (50.5/47.8) mAP. 
We tend to prefer the network with a higher mAP value when the network achieves a real-time performance (>10 FPS).
In this work, we use DIR-BHRNet-32 as the default lightweight network by considering a balanced performance of GFLOPS and mAP.

\begin{figure}[!ht] \centering    
    \includegraphics[width=.45\textwidth,trim={5 9 5 5},clip]{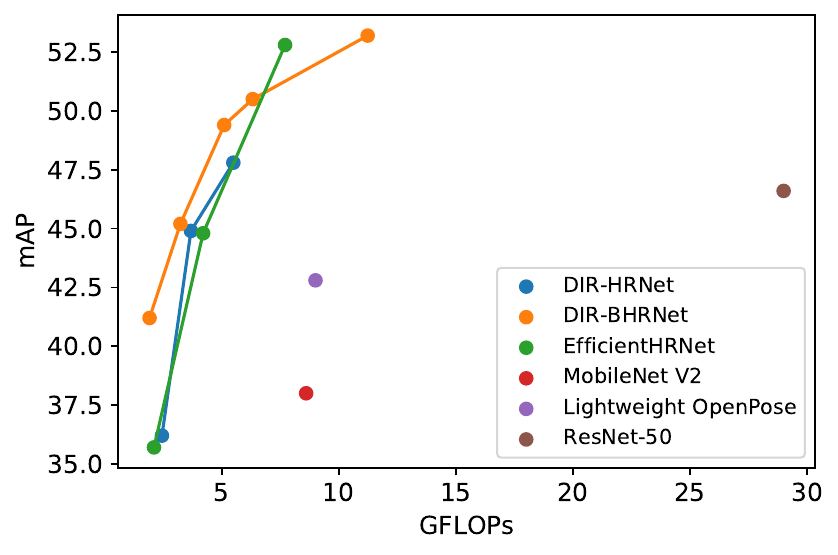} \vspace{-3pt}
    \caption{The comparison (Pareto) between our lightweight networks and the state-of-the-art real-time MPPE methods.} 
    \label{fig:comparison}
\end{figure} 

The higher mAP value indicates better performance and the lower GFLOPs indicate lower computational cost/latency. 
In this work, we expect a high value of mAP with low GFLOPs, which is a Pareto efficiency problem.
Therefore, we visually compare the performance of our lightweight networks with the state-of-the-art methods, as shown in \autoref{fig:comparison}.
The results show that DIR-BHRNet basically occupies the upper left, which means that DIR-BHRNet dominates the other methods.
Exceptionally, EfficientHRNet performs a 52.8 mAP with 7.7 GFLOPs. 
However, the network with 7.7 GFLOPs hardly performs real-time MPPE on mainstream smartphones.
In contrast, our DIR-BHRNet performs a balance performance of 50.5 mAP and 6.32 GFLOPs, which performs more than 10 FPS on mainstream smartphones (see \autoref{subsec:smartphones}).

\vspace{-9pt}
\subsection{Results on Smartphones}
\label{subsec:smartphones}
\vspace{-3pt}

\begin{figure}[!ht] \centering \small
    \begin{tabular}{c @{} c} 
         \includegraphics[width=0.18\textwidth, trim={125 25 375 5}, clip]{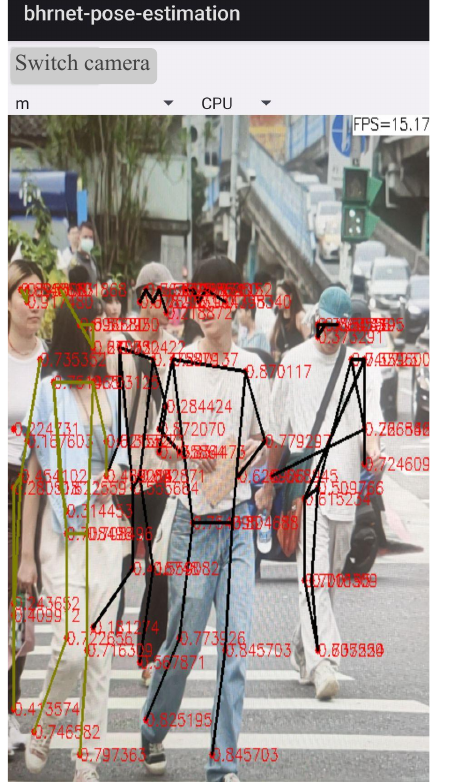} & 
         \includegraphics[width=0.28\textwidth, trim={1075 360 5500 430}, clip]{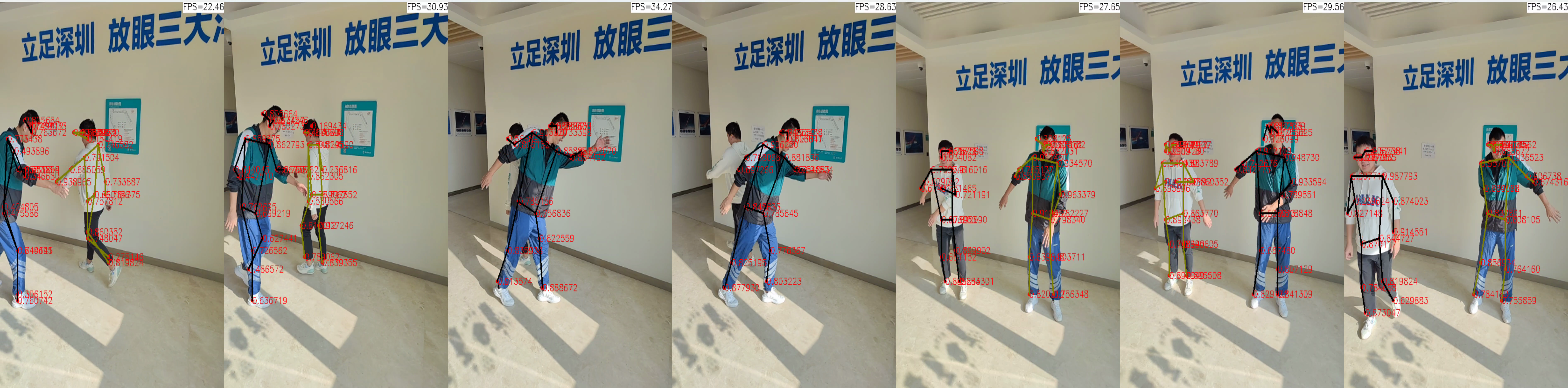} 
    \end{tabular} \vspace{-3pt}
    \caption{Visualization of the Android UI on smartphones with the DIR-BHRNet for real-time MPPE. For more details, please refer to the video description \textsuperscript{\ref{note1}}.}
    \label{fig:app}
\end{figure}

In this work, we integrate the trained DIR-HRNet and DIR-BHRNet into an Android executable file that runs on smartphones, as the framework in \autoref{fig:expeirments}.
The program collects images from the camera of smartphones as the input of DIR-BHRNet.
The visualization of the program with DIR-BHRNet on Android smartphones is shown in \autoref{fig:app}.
We evaluate the proposed methods, DIR-HRNet and DIR-BHRNet, from the aspects of latency and memory usage on smartphones.

 \vspace{-5pt}

\subsubsection{Memory Usage}
Memory usage is a crucial factor for real-time applications on mobile devices. 
The memory usage of a network in a program is approximately proportional to the number of network parameters.
\begin{table}[!ht] \centering \small \renewcommand{\arraystretch}{0.8} \setlength\tabcolsep{3pt} 
    \begin{tabular}{l c c l l} \toprule
    \multirow{2}{*}{Programs} & \multirow{2}{*}{Input size} & \multirow{2}{*}{Memory usage} & \multicolumn{2}{c} {Percentage} \\ \cline{4-5}
    & & & /8G (\%) & /12G (\%)  \\ \midrule
    \multirow{3}{*}{DIR-HRNet} & $256\times 256$ & 80MB & 1.0\% & 0.67\% \\
    & \cellcolor{gray!30} $384\times 384$ & \cellcolor{gray!30}112MB & \cellcolor{gray!30}1.4\% & \cellcolor{gray!30}0.93\% \\
    & $512\times 512$ & 148MB & 1.85\% & 1.23\% \\ \midrule
    \multirow{3}{*}{DIR-BHRNet} & $256\times 256$ & 90MB & 1.13\% & 0.75\% \\
    & \cellcolor{gray!30}$384\times 384$ & \cellcolor{gray!30}125MB & \cellcolor{gray!30}1.56\% & \cellcolor{gray!30}1.04 \% \\
    & $512\times 512$ & 172MB & 2.1\% & 1.43 \% \\  \bottomrule
    \end{tabular} \vspace{-3pt}
    \caption{Memory usage of our MPPE methods on Android smartphones which are observed in Android Studio.}
    \label{tab:mem_usage}
\end{table} \vspace{-3pt}
We therefore observe the memory usage of DIR-HRNet and DIR-BHRNet programs on Android smartphones rather than all other state-of-the-art networks.
Specifically, we collect the memory usage by using CPU Profiler, as shown in \autoref{tab:mem_usage}.
The memory usage of DIR-HRNet and DIR-BHRNet programs with 
\begin{table}[!ht]
    \centering \small  \setlength\tabcolsep{2pt} \renewcommand{\arraystretch}{0.8}
    \begin{tabular}{l l c c c} \toprule
    \makecell[l]{Networks \\ (input size)} & Smartphones & \#Params & {\footnotesize GFLOPS} & FPS(Latency/ms) \\ \midrule
    \multirow{3}{*}{\makecell{DIR-HRNet \\ $(256\times256)$}} & Redmi K40 &  \multirow{3}{*}{2.51M} & \multirow{3}{*}{2.44} & 24.7 / (40.5) \\
    & Xiaomi 10 &  &  & 29.2 / (34.2) \\
    & Xiaomi 12 &  &  & 32.3 / (31.0) \\ \midrule
    \multirow{3}{*}{\makecell{DIR-BHRNet \\ $(256\times256)$}} & Redmi K40 &  \multirow{3}{*}{6.03M} & \multirow{3}{*}{2.81} & 21.4 / (46.7) \\
    & Xiaomi 10 & &  &  25.3 / (39.5) \\
    & Xiaomi 12 &  &  27.9 / (35.8) \\ \midrule
    \multirow{3}{*}{\makecell{DIR-HRNet \\ $(384\times384)$}} & Redmi K40 & & & 10.8 / (92.6) \\
    & Xiaomi 10 & 2.51M & 5.49 & 13.2 / (75.8) \\
    & Xiaomi 12 &  &  & 13.7 / (73.0) \\ \midrule
    \rowcolor{gray!30} & Redmi K40 &  &  & \cellcolor{gray!30}\textbf{10.5 / (95.2)} \\
    \rowcolor{gray!30}  & Xiaomi 10 & 6.03M & 6.32 & \cellcolor{gray!30}\textbf{11.4 / (87.7)} \\
    \rowcolor{gray!30} \multirow{-3}{*}{\makecell{DIR-BHRNet \\ $(384\times384)$}} & Xiaomi 12 &  &  & \textbf{11.8 / (84.7)} \\ \midrule
    \multirow{3}{*}{\makecell{DIR-HRNet \\$(512\times512)$}} & Redmi K40 & \multirow{3}{*}{2.51M} & \multirow{3}{*}{9.77} & 4.0 / (250.0) \\
    & Xiaomi 10 &  &  & 5.1 / (196.1) \\
    & Xiaomi 12 &  &  & 5.4 / (185.2) \\ \midrule
    \multirow{3}{*}{\makecell{DIR-BHRNet \\ $(512\times512)$}}
    & Redmi K40 & \multirow{3}{*}{6.03M} & \multirow{3}{*}{11.23} & 3.4 / (294.1) \\
    & Xiaomi 10 &  &  & 4.1 / (244.0) \\
    & Xiaomi 12 &  &  & 4.3 / (232.6) \\ \bottomrule
    \end{tabular} \vspace{-3pt}
    \caption{The FPS and latency results of our networks with various input sizes on Android smartphones.}
    \label{tab:latency}
\end{table}
different input sizes ranges from 80MB to 172MB, which is from $0.67\%$ to $2.1\%$ of the used smartphones with 8G or 12G memory usage.
The results demonstrate that the memory usage of our lightweight model is reasonable for real-time applications on the current mainstream smartphones.


\subsubsection{Latency} 
To evaluate the latency of our methods, we test DIR-HRNet and DIR-BHRNet on Android smartphones with the input sizes of $256$$\times$$256$, $384$$\times$$384$ and $512$$\times$$512$.
We measure the frames per second of these methods on Android smartphones, as the results in \autoref{tab:latency}.
The DIR-HRNet with an input size of $384$$\times$$384$ performs 10.8 FPS, 13.2 FPS, and 13.7 FPS on the Android smartphones Redmi K40, Xiaomi 10, and Xiaomi 12, respectively.
For the same setup, DIR-BHRNet performs a reasonable speed of 10.5 FPS, 11.4 FPS, and 11.8 FPS for real-time MPPE.
In summary, DIR-HRNet and DIR-BHRNet perform real-time MPPE with a speed of more than 10 FPS on the current mainstream smartphones.

\vspace{-5pt}
\section{Conclusions}
\label{sec:conclusion}
In this work, we propose a novel lightweight network, DIR-BHRNet, for real-time multi-person pose estimation on smartphones. 
Specifically, we design a new convolution module, DIR, by adding an extra depthwise convolution and shortcut connection to enhance the spatial feature for improving accuracy.
Furthermore, we propose a novel network structure, BHRNet, to balance the distribution of computational costs and reduce computational costs by reconfiguring the number of convolutional blocks.
We integrate the DIR module and BHRNet structure to build the novel lightweight network, DIR-BHRNet, for real-time MPPE.

We investigate the performance of the proposed DIR module and the lightweight DIR-BHRNet on the well-known COCO and CrowdPose datasets.
The results demonstrate that the novel DIR module and BHRNet structure in DIR-BHRNet contribute to a higher accuracy with a similar computational cost than the state-of-the-art methods.
Furthermore, we implement the lightweight DIR-BHRNet on mainstream Android smartphones for real-time MPPE with a speed of more than 10 FPS.
Finally, we release the Android executable file and source code to facilitate the development of real-time MPPE on smartphones.
In the future, we will investigate the attention mechanism in lightweight networks for improving the accuracy of real-time MPPE.
Furthermore, we will extend the proposed lightweight networks for real-time 3D MPPE and implement these methods on smartphones.

\vspace{-3pt}
\bibliographystyle{IEEEtran}
\vspace{-3pt}
\bibliography{bibliography}

\begin{thebibliography}{10}
\providecommand{\url}[1]{#1}
\csname url@samestyle\endcsname
\providecommand{\newblock}{\relax}
\providecommand{\bibinfo}[2]{#2}
\providecommand{\BIBentrySTDinterwordspacing}{\spaceskip=0pt\relax}
\providecommand{\BIBentryALTinterwordstretchfactor}{4}
\providecommand{\BIBentryALTinterwordspacing}{\spaceskip=\fontdimen2\font plus
\BIBentryALTinterwordstretchfactor\fontdimen3\font minus \fontdimen4\font\relax}
\providecommand{\BIBforeignlanguage}[2]{{%
\expandafter\ifx\csname l@#1\endcsname\relax
\typeout{** WARNING: IEEEtran.bst: No hyphenation pattern has been}%
\typeout{** loaded for the language `#1'. Using the pattern for}%
\typeout{** the default language instead.}%
\else
\language=\csname l@#1\endcsname
\fi
#2}}
\providecommand{\BIBdecl}{\relax}
\BIBdecl

\bibitem{Lan2023vision}
G.~Lan, Y.~Wu, F.~Hu, and Q.~Hao, ``Vision-based human pose estimation via deep learning: A survey,'' \emph{IEEE Transactions on Human-Machine Systems}, vol.~53, no.~1, pp. 253--268, 2023.

\bibitem{newell2017associative}
A.~Newell, Z.~Huang, and J.~Deng, ``Associative embedding: End-to-end learning for joint detection and grouping,'' in \emph{Advances in neural information processing systems}, 2017, pp. 2277--2287.

\bibitem{cao2019openpose}
Z.~Cao, G.~Hidalgo, T.~Simon, S.-E. Wei, and Y.~Sheikh, ``{OpenPose}: realtime multi-person {2D} pose estimation using part affinity fields,'' \emph{IEEE transactions on pattern analysis and machine intelligence}, vol.~43, no.~1, pp. 172--186, 2019.

\bibitem{sandler2018mobilenetv2}
M.~Sandler, A.~Howard, M.~Zhu, A.~Zhmoginov, and L.-C. Chen, ``{MobileNetV2}: Inverted residuals and linear bottlenecks,'' in \emph{Proceedings of the IEEE CVPR}, 2018, pp. 4510--4520.

\bibitem{zhang2018shufflenet}
X.~Zhang, X.~Zhou, M.~Lin, and J.~Sun, ``{ShuffleNet}: An extremely efficient convolutional neural network for mobile devices,'' in \emph{Proceedings of the IEEE CVPR}, 2018, pp. 6848--6856.

\bibitem{neff2020efficienthrnet}
C.~Neff, A.~Sheth, S.~Furgurson, and H.~Tabkhi, ``{EfficientHRNet}: Efficient scaling for lightweight high-resolution multi-person pose estimation,'' \emph{arXiv preprint arXiv:2007.08090}, 2020.

\bibitem{cheng2020higherhrnet}
B.~Cheng, B.~Xiao, J.~Wang, H.~Shi, T.~S. Huang, and L.~Zhang, ``{HigherHRNet}: Scale-aware representation learning for bottom-up human pose estimation,'' in \emph{Proceedings of the IEEE CVPR}, 2020, pp. 5386--5395.

\bibitem{fang2022alphapose}
H.-S. Fang, J.~Li, H.~Tang, C.~Xu, H.~Zhu, Y.~Xiu, Y.-L. Li, and C.~Lu, ``{AlphaPose}: Whole-body regional multi-person pose estimation and tracking in real-time,'' \emph{IEEE Transactions on Pattern Analysis and Machine Intelligence}, 2022.

\bibitem{osokin2018real}
D.~Osokin, ``Real-time {2D} multi-person pose estimation on {CPU}: Lightweight {OpenPose},'' \emph{arXiv preprint arXiv:1811.12004}, 2018.

\bibitem{yu2021lite}
C.~Yu, B.~Xiao, C.~Gao, L.~Yuan, L.~Zhang, N.~Sang, and J.~Wang, ``{Lite-HRNet}: A lightweight high-resolution network,'' \emph{arXiv preprint arXiv:2104.06403}, 2021.

\bibitem{xiao2018simple}
B.~Xiao, H.~Wu, and Y.~Wei, ``Simple baselines for human pose estimation and tracking,'' in \emph{Proceedings of the European conference on computer vision (ECCV)}, 2018, pp. 466--481.

\bibitem{zhuang2024fastervoxelpose}
Z.~Zhuang and Y.~Zhou, ``Fastervoxelpose+: Fast and accurate voxel-based 3d human pose estimation by depth-wise projection decay,'' in \emph{Asian Conference on Machine Learning}.\hskip 1em plus 0.5em minus 0.4em\relax PMLR, 2024, pp. 1763--1778.

\bibitem{tan2019efficientnet}
M.~Tan and Q.~Le, ``{EfficientNet}: Rethinking model scaling for convolutional neural networks,'' in \emph{International Conference on Machine Learning}.\hskip 1em plus 0.5em minus 0.4em\relax PMLR, 2019, pp. 6105--6114.

\bibitem{cai2021yolobile}
Y.~Cai, H.~Li, G.~Yuan, W.~Niu, Y.~Li, X.~Tang, B.~Ren, and Y.~Wang, ``Yolobile: Real-time object detection on mobile devices via compression-compilation co-design,'' in \emph{Proceedings of the AAAI conference on artificial intelligence}, vol.~35, no.~2, 2021, pp. 955--963.

\bibitem{he2017channel}
Y.~He, X.~Zhang, and J.~Sun, ``Channel pruning for accelerating very deep neural networks,'' in \emph{Proceedings of the IEEE international conference on computer vision}, 2017, pp. 1389--1397.

\bibitem{shen2021towards}
X.~Shen, G.~Yuan, W.~Niu, X.~Ma, J.~Guan, Z.~Li, B.~Ren, and Y.~Wang, ``Towards fast and accurate multi-person pose estimation on mobile devices,'' \emph{arXiv preprint arXiv:2106.15304}, 2021.

\bibitem{chen2020adabert}
D.~Chen, Y.~Li, M.~Qiu, Z.~Wang, B.~Li, B.~Ding, H.~Deng, W.~Lin, and J.~Zhou, ``Adabert: Task-adaptive bert compression with differentiable neural architecture search,'' \emph{arXiv preprint arXiv:2001.04246}, 2020.

\bibitem{huynh2017deepmon}
L.~N. Huynh, Y.~Lee, and R.~K. Balan, ``{DeepMon}: Mobile {GPU}-based deep learning framework for continuous vision applications,'' in \emph{Proceedings of the 15th Annual International Conference on Mobile Systems, Applications, and Services}, 2017, pp. 82--95.

\bibitem{xu2018deepcache}
M.~Xu, M.~Zhu, Y.~Liu, and X.~Liu, ``{DeepCache}: Principled cache for mobile deep vision,'' in \emph{Proceedings of the 24th Annual International Conference on Mobile Computing and Networking}, 2018, pp. 129--144.

\bibitem{kim2019mulayer}
Y.~Kim, J.~Kim, D.~Chae, D.~Kim, and J.~Kim, ``$\mu${L}ayer: Low latency on-device inference using cooperative single-layer acceleration and processor-friendly quantization,'' in \emph{Proceedings of the Fourteenth EuroSys Conference 2019}, 2019, pp. 1--15.

\bibitem{seo2023mdpose}
S.~Seo, J.~Yoo, J.~Hwang, and N.~Kwak, ``Mdpose: real-time multi-person pose estimation via mixture density model,'' in \emph{Uncertainty in Artificial Intelligence}.\hskip 1em plus 0.5em minus 0.4em\relax PMLR, 2023, pp. 1868--1878.

\bibitem{jiang2024yolo}
Y.~Jiang, K.~Yang, J.~Zhu, and L.~Qin, ``Yolo-rlepose: Improved yolo based on swin transformer and rle-oks loss for multi-person pose estimation,'' \emph{Electronics}, vol.~13, no.~3, p. 563, 2024.

\bibitem{jiang2023rtmpose}
T.~Jiang, P.~Lu, L.~Zhang, N.~Ma, R.~Han, C.~Lyu, Y.~Li, and K.~Chen, ``Rtmpose: Real-time multi-person pose estimation based on mmpose,'' \emph{arXiv preprint arXiv:2303.07399}, 2023.

\bibitem{sun2019deep}
K.~Sun, B.~Xiao, D.~Liu, and J.~Wang, ``Deep high-resolution representation learning for human pose estimation,'' in \emph{Proceedings of the IEEE CVPR}, 2019, pp. 5693--5703.

\bibitem{lin2014microsoft}
T.-Y. Lin, M.~Maire, S.~Belongie, J.~Hays, P.~Perona, P.~Doll{\'a}r, and C.~L. Zitnick, ``Microsoft {COCO}: Common objects in context,'' in \emph{European conference on computer vision}.\hskip 1em plus 0.5em minus 0.4em\relax Springer, 2014, pp. 740--755.

\bibitem{li2019crowdpose}
J.~Li, C.~Wang, H.~Zhu, Y.~Mao, H.-S. Fang, and C.~Lu, ``{CrowdPose}: Efficient crowded scenes pose estimation and a new benchmark,'' in \emph{Proceedings of the IEEE CVPR}, 2019, pp. 10\,863--10\,872.

\bibitem{ncnn}
\BIBentryALTinterwordspacing
Tencent, ``{NCNN},'' 2017. [Online]. Available: \url{https://github.com/Tencent/ncnn}
\BIBentrySTDinterwordspacing

\bibitem{tobeta2022e2pose}
M.~Tobeta, Y.~Sawada, Z.~Zheng, S.~Takamuku, and N.~Natori, ``E2pose: Fully convolutional networks for end-to-end multi-person pose estimation,'' in \emph{2022 IEEE/RSJ International Conference on Intelligent Robots and Systems (IROS)}.\hskip 1em plus 0.5em minus 0.4em\relax IEEE, 2022, pp. 532--537.

\end{thebibliography}

\end{document}